\documentclass{llncs}

\usepackage{graphicx}
\usepackage{xspace}
\usepackage{listings}
\usepackage{hyperref}
\usepackage{subfig}
\usepackage{wrapfig}
\usepackage[T1]{fontenc}

\usepackage[colorinlistoftodos,textwidth=80pt]{todonotes}
\presetkeys{todonotes}{color=blue!20, bordercolor=blue!80, inline}{}

\newcommand{\mm}[1]{\mbox{${\rm MM}(#1)$}}

\newcommand{\embasp}{\protect\textsc{embASP}\xspace}

\newcommand{\dlvfit}{\emph{DLVfit}\xspace}

\newcommand{\derives}{\ensuremath{\mbox{\,$:$--}\,}\xspace}
\newcommand{\Or}{\ensuremath{|}\xspace}
\newcommand{\naf}{\ensuremath{\mathtt{not\ }}}

\newcommand{\dlv}{{\small DLV}\xspace}
\newcommand{\R}{\ensuremath{r}}

\newcommand{\p}{\ensuremath{{\cal P}}\xspace}

\newcommand{\BP}{\ensuremath{B_{\p}}\xspace}
\newcommand{\UP}{\ensuremath{U_{\p}}\xspace}
\newcommand{\BpR}{\ensuremath{B^+(\R)}}
\newcommand{\BnR}{\ensuremath{B^-(\R)}}

\newcommand{\Core}{\textit{Core}\xspace}
\newcommand{\Languages}{\textit{ASP Language}\xspace}
\newcommand{\Platforms}{\textit{Platforms}\xspace}
\newcommand{\Systems}{\textit{Systems}\xspace}
\newcommand{\Framework}{\textit{Framework}\xspace}
\newcommand{\Handler}{\textit{Handler}\xspace}
\newcommand{\InputProgram}{\textit{Input Program}\xspace}
\newcommand{\OptionDescriptor}{\textit{Option Descriptor}\xspace}
\newcommand{\Service}{\textit{Service}\xspace}
\newcommand{\Callback}{\textit{Callback}\xspace}
\newcommand{\Output}{\textit{Output}\xspace}

\usepackage{hyphenat}
\newenvironment{dlvcode}
  {\begin{displaymath}\begin{array}{l}}
  {\end{array}\end{displaymath}}

\begin{document}

\mainmatter
\title{A Framework for Easing the Development of Applications Embedding Answer Set Programming\thanks{ Original work published in http://dl.acm.org/citation.cfm?id=2968594}}

\author{Francesco Calimeri\inst{1,2}
 \and Davide Fusc\`{a}\inst{1} \and Stefano Germano\inst{1} \and \\
Simona Perri\inst{1} \and Jessica Zangari\inst{1}}

%\authorrunning{F. Calimeri et al.}
%\titlerunning{Boosting the Development of ASP-based Applications in Mobile and General Scenarios}

\institute{
	Department of Mathematics and Computer Science, University of Calabria, Italy \\
	\email{$\{$calimeri,fusca,germano,
perri,zangari$\}$@mat.unical.it}
\and
    DLVSystem Srl, Italy\\
	\email{calimeri@dlvsystem.com}
}

\maketitle
\begin{abstract}

Answer Set Programming (ASP) is a well-established declarative problem solving paradigm which became widely used in AI and recognized as a powerful tool for knowledge representation and reasoning (KRR), especially for its high expressiveness and the ability to deal also with incomplete knowledge.

Recently, thanks to the availability of a number of robust and efficient implementations, ASP has been increasingly employed in a number of different domains, and used for the development of industrial-level and enterprise applications. This made clear the need for proper development tools and interoperability mechanisms for easing interaction and integration with external systems in the widest range of real-world scenarios, including mobile applications and educational contexts.

In this work we present a framework for integrating the KRR capabilities of ASP into generic applications. We show the use of the framework by illustrating proper specializations for some relevant ASP systems over different platforms, including the mobile setting; furthermore, the potential of the framework for educational purposes is illustrated by means of the development of several ASP-based applications.
\end{abstract}

\keywords{Answer Set Programming; Logic Programs; Education; Industrial Applications; Knowledge Representation and Reasoning; Object-Oriented Programming; Software Development; Complex Systems; Embedded Systems; Artificial Intelligence}

\section{Introduction}
Answer Set Programming (ASP)~\cite{bara-2002,brew-etal-2011-cacm,eiter2000declarative,eite-etal-2009-primer,gelf-lifs-91,mare-trus-99,niem-99} is a purely declarative formalism for knowledge representation and reasoning developed in the field of logic programming and nonmonotonic reasoning.
The language of ASP is based on rules, allowing (in general)
for both disjunction in rule heads and nonmonotonic negation in the body.

The idea of answer set programming is to represent a given computational problem by the means of a logic program whose intended models, called \emph{answer sets}, correspond one-to-one to solutions; hence, an answer set solver can be used in order to actually find such solutions~\cite{lifs-99a}.
The term ``Answer Set Programming''
was introduced by Vladimir Lifschitz to denote a declarative programming methodology~\cite{lifs-99a}; concerning
terminology, ASP is sometimes used in a somewhat broader sense,
referring to any declarative formalism which represents
solutions as sets.
However, the more frequent understanding is the one adopted in
this article, which dates back to \cite{gelf-lifs-91}.
For introductory material on ASP, we refer to
\cite{bara-2002,gelf-leon-02,lifs-99a,mare-trus-99}.

\medskip

After more than twenty years of research, the theoretical properties of ASP are well understood and the solving technology, as evidenced by the availability of a number of robust and efficient systems~\cite{calimeri2016design}, is mature for practical applications: ASP has been increasingly employed in many different domains, and also used for the development of industrial-level and enterprise applications~\cite{calimeri2013application,leon-ricc-2015-rw-invited}. Notably, this is spreading ASP teaching in universities worldwide, and, interestingly, is moving the focus from a strict theoretical scope to more practical aspects. Moreover, it makes clear the need for proper tools and interoperability mechanisms that ease the development of ASP-based applications, in both educational and real-world contexts.

\medskip

In this work, we present a framework for the integration of ASP in external systems for generic applications; it consists of an abstract architecture, implementable in a programming language of choice, that easily allows for proper specializations to different platforms and ASP reasoners.

The framework features explicit mechanisms for two-way translations between strings
recognizable by ASP solvers and objects in the programming language at hand, directly
employable within applications. This gives developers the possibility to work separately on ASP-based modules and on applications that makes use of them, and keeps things simple when developing complex applications. Let us think, for instance, of a scenario in which different figures are involved, such as Android/Java developers and KRR experts. Both figures can take advantage from the fact that the knowledge base and the reasoning modules can be designed and developed independently from the rest of the Java-based application.

\medskip

In order to illustrate the use of the framework, we present here an actual Java implementation; in addition, we introduce two specialized libraries for DLV~\cite{leon-etal-2002-dlv} and clingo~\cite{gekakasc14b}, two state-of-the-art ASP systems, on mobile and desktop platforms, respectively. Furthermore, we show some applications developed in an educational context, that prove the effectiveness of the framework.

\section{Answer Set Programming}\label{sec:preliminaries}\label{sec:dlp}\label{sec:ASP}
In this section, we briefly recall syntax and semantics of Answer Set Programming.

\medskip

It is worth recalling that a significant amount of work has been carried out by the scientific community for extending the basic language, in order to increase the expressive power and improve usability of the formalism. This has led to a variety of ASP ``dialects'', supported by a corresponding variety of ASP systems, that only share a portion of the basic language. Notably, the community recently agreed on the definition of a standard input language for ASP systems, namely ASP-Core-2~\cite{calimeri2012asp}, which is also the official language of the ASP Competition series~\cite{DBLP:conf/aaai/GebserMR16}; it features most  of the advanced constructs and mechanisms with a well-defined semantics that have been introduced and implemented in the latest years.

\medskip

For the sake of simplicity, we focus next on the basic aspects of the language; for a complete reference to the ASP-Core-2 standard, and further details about advanced ASP features, we refer the reader to~\cite{calimeri2012asp} and the vast literature.

\subsection{Syntax}
A variable or a constant is a {\em term}.
An {\em atom} is
$a(t_{1}, \dots,$ $t_{n})$, where $a$ is a {\em predicate} of arity $n$
and $t_{1}, \dots, t_{n}$ are terms.  A {\em literal} is either a
{\em positive~literal} $p$ or a {\em negative~literal} $\naf p$,
where $p$ is an atom. A {\em disjunctive rule} ({\em rule}, for
short) \R{} is a formula

\medskip

$
a_1\ \Or\ \cdots\ \Or\ a_n\ \derives\
        b_1,\cdots, b_k,\
        \naf\ b_{k+1},\cdots,\ \naf\ b_m.
$

\medskip

where $a_1,\cdots ,a_n,b_1,\cdots ,b_m$ are atoms and $n\geq 0,$
$m\geq k\geq 0$.  The disjunction $a_1\ \Or\ \cdots\ \Or\ a_n$ is
the {\em head} of \R{}, while the conjunction $b_1 , ..., b_k,\
\naf\ b_{k+1} , ...,\ \naf\ b_m$ is the {\em body} of \R{}. A rule
without head literals (i.e.\ $n=0$) is usually referred to as an
{\em integrity constraint}.
If the body is empty (i.e.\ $k=m=0$), it is called a {\em fact}.

$H(r)$ denotes the set $\{a_1 ,..., a_n \}$ of the head atoms, and
by $B(r)$ the set $\{b_1 ,..., b_k,$ $\naf b_{k+1} ,\ldots , \naf
b_m \}$ of the body literals. $\BpR$ (resp., $\BnR$) denotes the
set of atoms occurring positively (resp., negatively) in $B(r)$. A
rule $r$ is {\em safe} if each variable appearing in $r$ appears
also in some positive body literal of $r$.

An {\em ASP program } $\p$ is a finite set of safe rules.
An atom, a literal, a
rule, or a program is {\em ground} if no variables appear in it.
Accordingly with the database terminology, a predicate occurring
only in {\em facts} is referred to as an {\em EDB} predicate, all
others as {\em IDB} predicates; the set of facts of $\p$ is
denoted by $EDB(\p)$.

\subsection{Semantics}\label{subsec:semantics}
Let $\p$ be a program. The {\em Herbrand Universe} and the {\em
Herbrand Base} of $\p$ are defined in the standard way and denoted
by $\UP$ and $\BP$, respectively.

Given a rule $r$ occurring in $\p$, a {\em ground instance} of $r$
is a rule obtained from $r$ by replacing every variable $X$ in $r$
by $\sigma (X)$, where $\sigma$ is a substitution mapping the
variables occurring in $r$ to constants in $\UP$; $ground( \p)$
denotes the set of all the ground instances of the rules occurring
in $\p$.

An {\em interpretation} for $\p$ is a set of ground atoms, that
is, an interpretation is a subset $I$ of $\BP$. A ground positive
literal $A$ is {\em true} (resp., {\em false}) w.r.t. $I$ if $A
\in I$ (resp., $A \not\in I$). A ground negative literal $\naf A$
is {\em true} w.r.t. $I$ if $A$ is false w.r.t. $I$; otherwise
$\naf A$ is false w.r.t. $I$.
Let $r$ be a ground rule in $ground( \p )$.  The head of $r$ is
{\em true} w.r.t. $I$ if $H(r) \cap I \neq \emptyset$.  The body
of $r$ is {\em true} w.r.t. $I$ if all body literals of $r$ are
true w.r.t. $I$ (i.e., $B^+(r) \subseteq I$ and $B^-(r)\cap I =
\emptyset$) and is {\em false} w.r.t. $I$ otherwise.  The rule $r$
is {\em satisfied} (or {\em true}) w.r.t.  $I$ if its head is true
w.r.t. $I$ or its body is false w.r.t. $I$.
A {\em model} for $\p$ is an interpretation $M$ for $\p$ such that
every rule $r \in ground(\p)$ is true w.r.t. $M$.  A model $M$ for
$\p$ is {\em minimal} if no model $N$ for $\p$ exists such that
$N$ is a proper subset of $M$.  The set of all minimal models for
$\p$ is denoted by ${\rm MM}(\p )$.

Given a ground program  $\p$ and an interpretation  $I$, the {\em
reduct} of $\p$ w.r.t. $I$ is the subset $\p^I$ of $\p$, which is
obtained from $\p$ by deleting rules in which a body literal is
false w.r.t. $I$.
Note that the above definition of reduct, proposed in
\cite{fabe-etal-2004-jelia}, simplifies the original definition of
Gelfond-Lifschitz (GL) transform \cite{gelf-lifs-91}, but is fully
equivalent to the GL transform for the definition of answer sets
\cite{fabe-etal-2004-jelia}.

  Let  $I$ be an interpretation for a
program  $\p$.  $I$ is an {\em answer set} (or stable model) for  $\p$ if $I \in
\mm{\p^I}$ (i.e.,  $I$ is a minimal model for the program
 $\p^I$) \cite{przy-91,gelf-lifs-91}. The
set of all answer sets for $\p$ is denoted by  $ANS(\p)$.

\subsection{Knowledge Representation and Reasoning with ASP}\label{subsec:KRR}
In the following, we briefly introduce the use of ASP as a tool for knowledge representation and reasoning, and show how its fully declarative nature allows to encode a large variety of problems via simple and elegant logic programs.

The examples below have been implemented adhering to the ``Guess\&Check'' ($GC$) paradigm~\cite{eiter2000declarative}, one of the most common ASP programming methodology.
In summary, a $GC$ program features 2 modules:
\begin{itemize}
\item a \textbf{Guessing Part}, that defines the search space (for instance, by means of disjunctive rules);
\item a \textbf{Checking Part}(optional), that checks solution admissibility (usually, by means of integrity constraints).
\end{itemize}
When dealing with optimization problems, the methodology can be further extended to match a ``Guess/Check/Optimize''\cite{bucc-etal-97a} ($GCO$) paradigm: ad-hoc means for expressing preferences among answer sets are employed, such as \emph{weak constraints}\cite{bucc-etal-97a,calimeri2012asp}, thus implementing the
\begin{itemize}
\item \textbf{Optimizing Part} (optional), that specifies preference criteria.
\end{itemize}

\medskip

\textbf{[3-COL]} As a first example, let us consider the well-known problem of 3-colorability, which consists of the assignment of three colors to the nodes of a graph in such a way that adjacent nodes always have different colors. This problem is known to be NP-complete.

Suppose that the nodes and the arcs are represented by a set $F$ of facts with
predicates $node$ (unary) and $arc$ (binary), respectively. Then, the
following ASP program allows us to determine the admissible ways of coloring
the given graph.

\begin{dlvcode}
r_1:\ \ color(X,r)\ \Or\ color(X,y)\ \Or\ color(X,g)
\derives node(X).\\
r_2:\ \ \derives arc(X,Y), color(X,C), color(Y,C).
\end{dlvcode}

Rule $r_1$ (\emph{guess}) above states that every node of the graph must be colored as {\bf r}ed or {\bf y}ellow or {\bf g}reen; $r_2$ (\emph{check}) forbids the assignment of the same color to any couple of adjacent nodes. The minimality of answer sets guarantees that every node is assigned only one color. Thus, there is a one-to-one correspondence between the solutions of the 3-coloring problem for the instance at hand and the answer sets of $F\cup\{r_1, r_2\}$: the graph represented by $F$ is 3-colorable if and only if $F\cup\{r_1, r_2\}$ has some answer set.

\medskip

We have shown how it is possible to deal with a problem by means of an ASP program such that the instance at hand has some solution if and only if the ASP program as some answer set; in the following, we show an ASP program whose answer sets witness that a property does not hold, i.e., the property at hand holds if and only if the program has no answer sets.

\medskip

\textbf{[RAMSEY]} The Ramsey Number $R(k,m)$ is the least integer $n$ such that, no matter how we color the arcs of the complete graph (clique) with $n$ nodes using two colors, say red and blue, there is a red clique with $k$ nodes (a red $k$-clique) or a blue clique with $m$ nodes (a blue $m$-clique). Ramsey numbers exist for all pairs of positive integers $k$ and $m$ \cite{radz-94}.

Similarly to what already described above, let $F$ be the collection of facts for input predicates {\em node} (unary) and {\em edge} (binary), encoding a complete graph with $n$ nodes; then, the following ASP program $P_{R(3,4)}$ allows to determine whether a given integer $n$ is the Ramsey Number $R(3,4)$, knowing that no integer smaller than $n$ is $R(3,4)$.

\begin{dlvcode}
r_1:\ \ blue(X,Y)\ \Or\ red(X,Y) \derives\ edge(X,Y). \\
 \ \\
r_2:\ \ \derives\ red(X,Y), red(X,Z), red(Y,Z). \\
r_3:\ \ \derives\ blue(X,Y), blue(X,Z), blue(Y,Z),  \\
 \ \ \ \ \ \ \ \ \ \ \ \ blue(X,W), blue(Y,W), blue(Z,W).
\end{dlvcode}

Intuitively, the disjunctive rule $r_1$ guesses a color for each edge. The first constraint $r_2$ eliminates the colorings containing a red complete graph (i.e., a clique) on 3 nodes; the second constraint $r_3$ eliminates the colorings containing a blue clique on 4 nodes. The program $P_{R(3,4)} \cup F$ has an answer set if and only if there is a coloring of the edges of the complete graph on $n$ nodes containing no red clique of size 3 and no blue clique of size 4. Thus, if there is an answer set for a particular $n$, then $n$ is \underline{not} $R(3,4)$, that is, $n < R(3,4)$. The smallest $n$ such that no answer set is found is the Ramsey Number $R(3,4)$.

\medskip

Eventually, let us show how ASP can be applied for solving puzzles.

\medskip

\textbf{[SUDOKU]} A classic Sudoku puzzle consists of a tableau featuring 81 cells, or positions, arranged in a 9*9 grid, which is divided into nine sub-tableaux (regions, or blocks) containing nine positions each. Initially, a number of positions (between 17 and 35) are filled with a number picked up in the range $1\dots 9$. The aim of the game is to check whether every empty position can be filled with a number between 1 and 9 in such a way that each row, column and block show all digits from 1 to 9 exactly once.

Let us suppose that a set of facts $F$ is given, representing the schema to be completed; in particular, a binary predicate {\em pos} encodes possible position coordinates; {\em symbol} is a unary predicate encoding possible symbols (numbers); facts of the form {\em sameblock(x1,y1,x2,y2)} state that two positions $(x1, y1)$ and $(x2, y2)$ are within the same block; facts of the form {\em cell(x,y,n)} represent that a position $(x,y)$ is filled with symbol $n$.

\begin{figure*}[t]
	\begin{center}
		\includegraphics[width=0.8\linewidth]{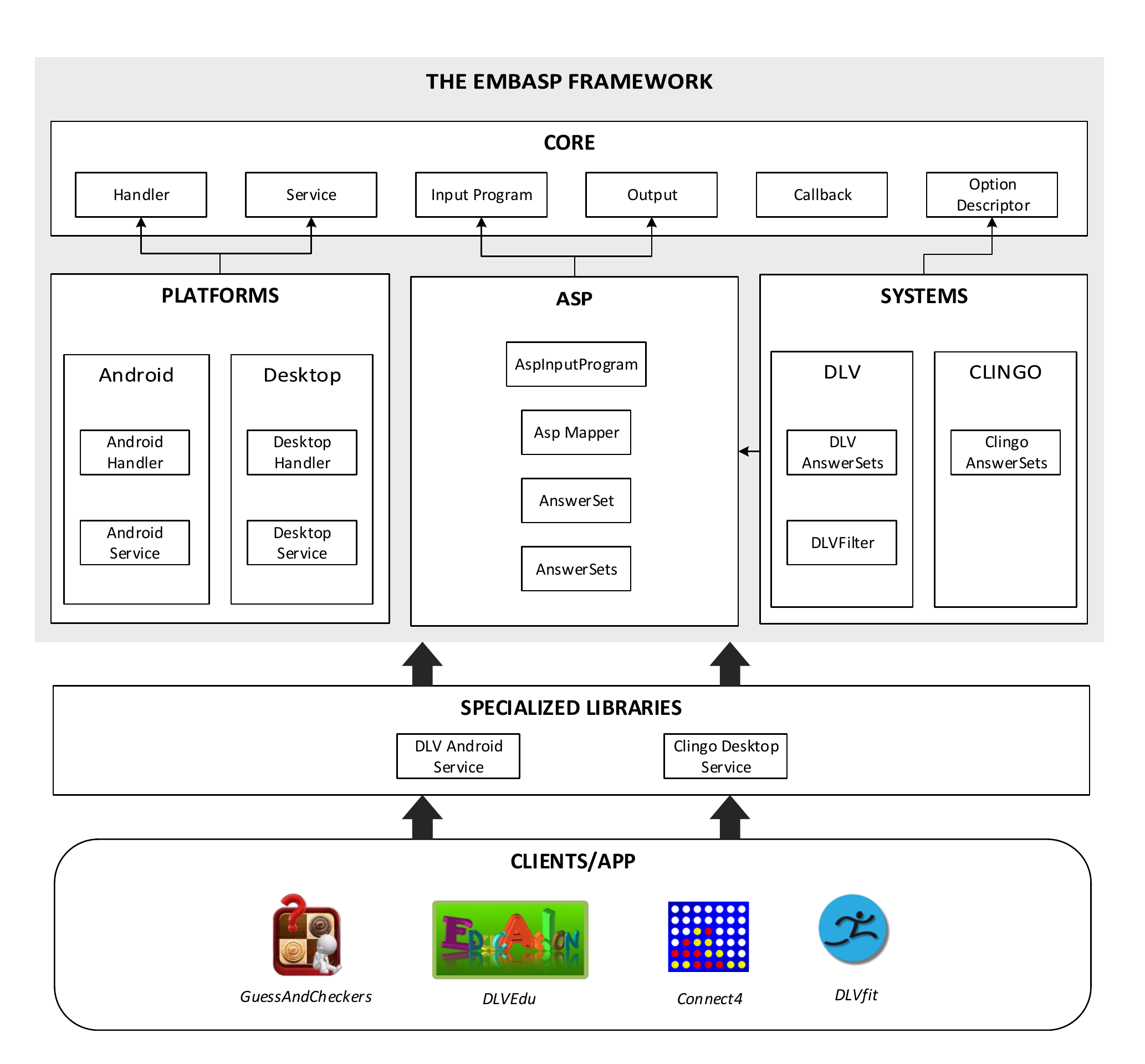}
		\caption{A visual overview of \embasp: abstract \Framework, some possible \textit{Specialized Libraries}, some examples of ASP-based applications relying on such libraries.}
		\label{fig:architecture}
	\end{center}
\end{figure*}

We show next an ASP program $P_{sudoku}$ such that the answer sets of $P_{sudoku} \cup F$ correspond to the solutions of the Sudoku schema at hand; note that, in general, well-founded sudoku instances have only one solution, and thus $P_{sudoku} \cup F$ will have a single answer set.

\begin{dlvcode}

r_1:\ \ \ cell(X,Y,N)\ \Or\ nocell(X,Y,N) \derives\ pos(X), \\
 \ \ \ \ \ \ \ \ pos(Y), symbol(N).\\
\ \\
r_2:\ \ \ \derives\ cell(X,Y,N), cell(X,Y,N1), N1 <> N. \\
\ \\
r_3:\ \ \ assigned(X,Y) \derives\ cell(X,Y,N).\\
r_4:\ \ \ \derives\ pos(X),pos(Y), not\ assigned(X,Y).\\
\ \\
r_5:\ \ \ \derives\ cell(X,Y1,Z), cell(X,Y2,Z), Y1<>Y2.\\
r_6:\ \ \ \derives\ cell(X1,Y,Z), cell(X2,Y,Z), X1<>X2.\\
 \ \\
r_{7}:\ \ \derives\ cell(X1,Y1,Z), cell(X2,Y2,Z), Y1 <> Y2, \\
\ \ \ \ \ \ \ \ \ \ \ sameblock(X1,Y1,X2,Y2).\\
r_{8}:\ \ \derives\ cell(X1,Y1,Z), cell(X2,Y2,Z), X1 <> X2, \\
\ \ \ \ \ \ \ \ \ \ \ sameblock(X1,Y1,X2,Y2).\\
\end{dlvcode}

Rules $r_{1}-r_{4}$ guess the number for each cell, ensuring that each cell is filled exactly one number (\textit{symbol}); note that the guessed values for the positions complete the extension of the predicate {\em cell} for which some values have been already provided in $F$. Rules $r_{5}$ and $r_{6}$ check that a number does not occur more than once in the same row or column, respectively; rules $r_{7}$ and $r_{8}$, finally, ensure that two different cells in the same block don't have the same number.

\section{The Framework}

In this section we introduce \embasp, an abstract framework for the integration of ASP in external systems for generic applications; then, we propose a Java implementation.

\begin{figure*}[t]
	\begin{center}
		\includegraphics[width=1.0\linewidth]{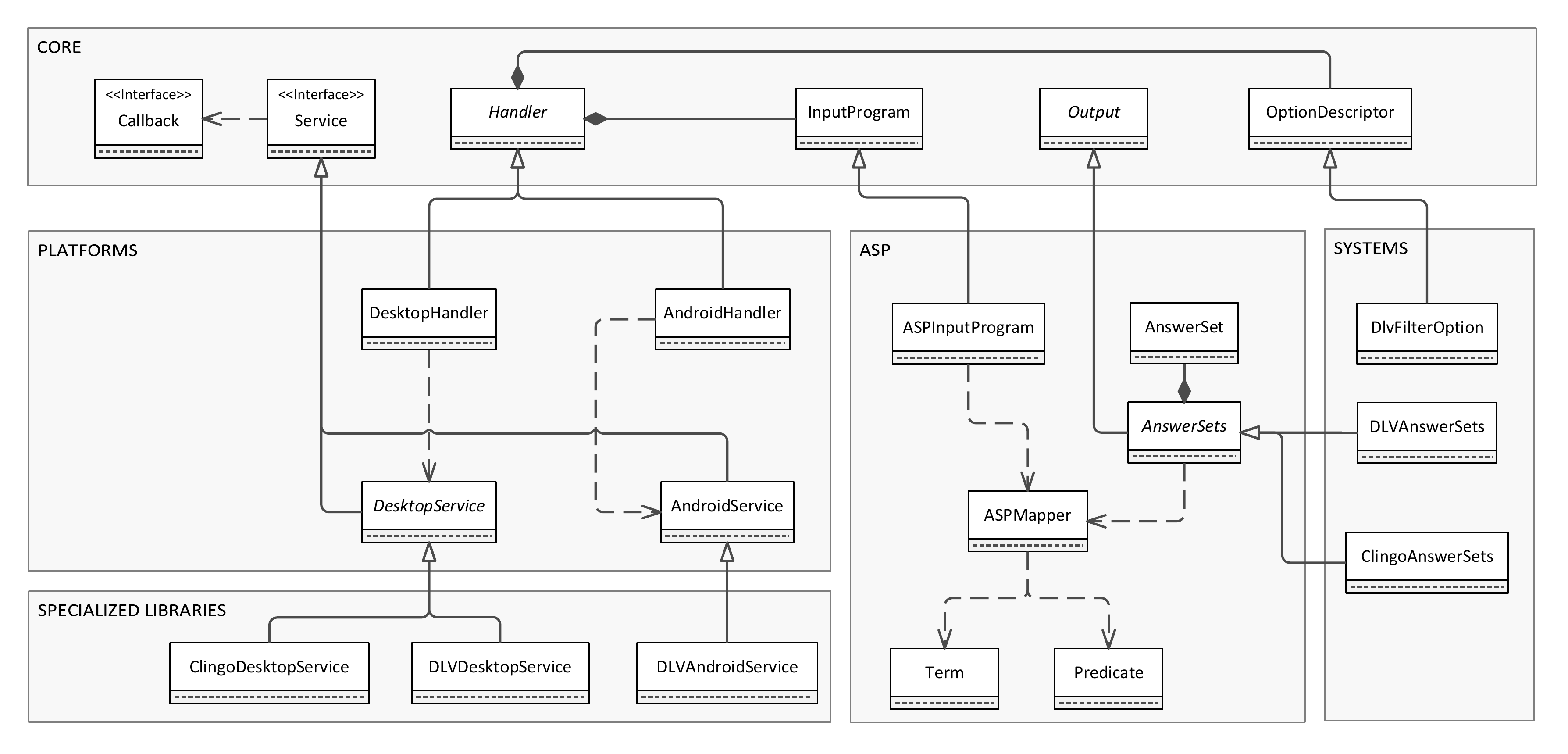}
		\caption{Simplified class diagram of the provided Java implementation of \embasp, then specialized to \dlv on Android and clingo on Desktop.}
		\label{fig:classdiagram}
	\end{center}
	\end{figure*}

The general architecture of \embasp is depicted in Figure~\ref{fig:architecture}
: it defines an abstract framework to be implemented in some object-oriented programming language.
Due to its abstract nature, Figure~\ref{fig:architecture} just reports the general dependencies among the main modules. Nevertheless, each concrete implementation might require specific dependencies among the inner components of each module, as can be observed in Figure~\ref{fig:classdiagram}, which is related to a concrete Java implementation and will be discussed hereafter.

\medskip

It is worth noting that the framework design is intended to ease and guide the generation of suitable libraries for the use of specific solvers on particular platforms; resulting applications manage ASP solvers as ``black boxes''.
On the one hand, this might lead to issues arising from users demanding for a more interactive white-box usage; on the other hand, this made us able to keep a clean design that grants an intuitive usage and an architecture which is general and easily adaptable to different platforms and reasoners.
The resulting libraries can hence be used in order to effectively embed ASP reasoning modules, handled by the ASP system(s) at hand, within any kind of application developed for the targeted platforms.
In addition, as already discussed above, the framework is meant to give developers the possibility to work separately on ASP-based modules and on the applications that makes use of them, thus keeping things simple when developing complex applications.
Additional specific advantages/disadvantages might arise depending on the programming language chosen for deploying libraries and on the target platform; special features, indeed, can make implementation, and in turn extensions and usage, easier or more difficult, to different extents.
We will briefly discuss these issues later on.

\subsection*{Abstract Architecture}

The framework architecture has been designed by means of four modules: \Core , \Platforms , \Languages, and \Systems, whose indented behaviour is described next.

\paragraph*{Core Module}
The \Core module defines the basic components of the \Framework.

The \Handler component mediates the communication between the \Framework and the user that can provide it with the input program(s) via the component \InputProgram, along with any desired solver's option(s) via the component \OptionDescriptor. A \Service component is meant for managing the chosen ASP solver executions.

Two different execution modes can be made available: synchronous or asynchronous. While in the synchronous mode any call to the execution of the ASP solver is \emph{blocking} (i.e., the caller waits until the reasoning task is completed), in asynchronous mode the call is non-blocking: a \Callback component notifies the caller once the reasoning task is completed. The result of the execution (i.e., the output of the ASP system) is handled by the \Output component, in both modes.

\paragraph*{Platforms Module}
The \Platforms module is meant for containing what is platform-dependent; in particular, the \Handler and \Service components from the \Core module that should be adapted according to the platform at hand, since they take care of practically launching solvers.

\paragraph*{ASP Language Module}
The \Languages module defines specific facilities for ASP; in particular, components \textit{AnswerSet} and \textit{AnswerSets} adapt \Output component to the ASP case.
Moreover, an additional component, namely the \textit{ASPMapper}, is conceived as an utility for managing input and output via objects, if the programming language at hand permits it.

\paragraph*{Systems Module}
The \Systems module defines what is system\hyp{}dependent; in particular, the \InputProgram, \Output and \OptionDescriptor components from the \Core module should be adapted in order to effectively interact with the ASP system at hand.

\subsection*{Implementing \embasp}
In the following, we propose a Java\footnote{\url{https://www.oracle.com/java}} implementation of the architecture described above, along with proper specializations for two of the state-of-the-art ASP systems. In particular, we implemented the main modules by means of classes or interfaces, and we created specialized libraries that permit the use of DLV on Android\footnote{\url{http://developer.android.com}} and clingo on desktop (i.e., any java-enabled desktop for which clingo is available).

Figure~\ref{fig:classdiagram} provides some details about classes and interfaces of the implementation.
For the sake of presentation, we do not report the complete UML~\cite{citeulike:1104855} class diagram, which is quite involved; rather, we illustrate a simplified version. Although methods inside classes have been omitted to further improve readability, adopted connectors follow UML syntax.
In order to better outline correspondences with the abstract architecture of Figure~\ref{fig:architecture}, classes belonging to a module have been grouped together. The complete UML class diagram is available online at \cite{embasp-web}.

\subsubsection*{Core module implementation}
Each component in the \Core module has been implemented by means of an homonymous class or interface.
In particular, the \texttt{Handler} class collects \texttt{InputProgram} and \texttt{OptionDescriptor} objects communicated by the user.

For what the asynchronous mode is concerned, the class \texttt{Service} depends from the interface \texttt{Callback}, since once the reasoning service has terminated, the result of the computation is returned back via a class \texttt{Callback}.

\subsubsection*{Platforms module implementation}
In order to support a new platform, the \Handler and \Service components must be adapted.

As for the Android platform, we developed an \texttt{AndroidHandler} that handles the execution of an \texttt{AndroidService}, which provides facilities to manage the execution of an ASP reasoner on the Android platform.

Similarly, for the desktop platform we developed a \texttt{DesktopHandler} and a \texttt{DesktopService}, which generalizes the usage of an ASP reasoner on the desktop platform, allowing both synchronous and asynchronous execution modes.

While both synchronous and asynchronous modes are provided in the desktop setting, we stick to the asynchronous one on Android: indeed, mobile users are familiar with apps featuring constantly reactive graphic interfaces, and according to this native asynchronous execution policy, we want to discourage a blocking execution.

\subsubsection*{ASP Language module implementation}
This module includes specific classes for the management of input and output to ASP solvers. In particular, \texttt{ASPInputProgram} extends \texttt{InputProgram} to the ASP case. In addition, since the ``result'' of an ASP solver execution consists of answer sets, the \Output class has been extended by the \texttt{AnswerSets} class that is composed by a set of \texttt{AnswerSet} objects.

Moreover, the module features an \texttt{ASPMapper} class, that acts like a translator, providing proper means for a two-way translation between strings recognizable by the ASP solver at hand and Java objects directly employable within the application. The \texttt{ASPMapper} is intended at translating ASP input and output from and to objects: thus has a dependency from \texttt{ASPInputProgram} and \texttt{AnswerSets} classes.

In our implementation, such translations are guided by Java Annotations\footnote{\url{https://docs.oracle.com/javase/tutorial/java/annotations/}},
a form of metadata that mark Java code and provide information that is not part of the program itself: they have no direct effect on the operation of the code they annotate. They have a number of uses, such as directions to the compiler, compile-time and deployment-time processing, or runtime processing. For more details, we refer the reader to the Java documentation.

In our setting, we make use of such feature so that it is possible to translate facts into strings and vice-versa via two custom annotations, defined according to the following syntax:
\begin{itemize}
\item
    \emph{@Predicate (string\_name)}: the target must be a class, and defines the predicate name the class is mapped to;
\item
    \emph{@Term (integer\_position)}: the target must be a field of a class annotated via @Predicate, and defines the term (and its position) in the ASP atom the field is mapped to.
\end{itemize}

By means of the Java Reflection mechanisms, annotations are examined at runtime, and taken into account to properly define the translation.

The user has to register all its annotated classes to the \texttt{ASPMapper}, although classes involved in input translation are automatically detected. If the classes intended for the translation are not annotated or not correctly annotated, an exception is raised. Other problems might occur if once that the solver output is returned, the user asks for a translation into objects of not annotated classes: in this case a warning is raised and the request is ignored.

Notably, such feature is meant to give developers the possibility to work separately on the ASP-based modules and on the Java side. The mapper acts like a middle-ware that enables the communication among the modules, and eases the burden of developers by means of an explicit, ready-made mapping between Java objects and the logic modules.

Further insights about this feature are illustrated thanks to an example in the next section.

\subsubsection*{Systems Module Implementation}
The classes \texttt{DLVAnswerSets} and \texttt{ClingoAnswerSets} implement specific extensions of the \texttt{AnswerSets} class, in charge of manipulating the output of the respective solvers.

Moreover, this module also contains classes extending \texttt{OptionDescriptor} to implement specific options of the solver at hand. For instance, the class \texttt{DLVFilter} is a utility class representing the filter option of DLV.

\subsection*{Specializing the Framework}
We implemented two libraries derived from \embasp, allowing the embedding of ASP reasoning modules handled by DLV and clingo, from within Android and desktop applications, respectively.

The classes \texttt{DLVAndroidService} and \texttt{ClingoDesktopService} are in charge of this task.

\texttt{DLVAndroidService} is a specific version of \texttt{AndroidService} for the execution of DLV on Android. It is worth noting that DLV was not available for Android; furthermore, it is natively implemented in C++, while the standard development process on Android is based on Java. To this end, DLV has been on purpose rebuilt using the NDK (Native Development Kit)\footnote{\url{https://developer.android.com/tools/sdk/ndk}}, and has been linked to the Java code using the JNI (Java Native Interface)\footnote{\url{http://docs.oracle.com/javase/8/docs/technotes/guides/jni}}. This grants the access to the APIs provided by the Android NDK, and in turn accedes to the \dlv exposed functionalities directly from the Java code of an Android application.

Similarly, \texttt{ClingoDesktopService} is a specific version tailored for the clingo reasoner on the desktop platform, extending the \texttt{DesktopService} with proper functions needed to invoke clingo. In this case, different versions of the solver for several desktop OSes were already available online~\cite{potasscoWebPage,gekakaosscsc11a}.

\section{Embedding ASP Programs}

In the following we show the use of the specialized Java libraries generated via \embasp for developing Android applications; we report some considerations about programming languages different from Java at the end of the section.

\medskip

As a use case, we will refer to an application for solving Sudoku puzzles.  We will report the code related to the \embasp usage; the complete code is available online~\cite{embasp-web}. Notably, thanks to the annotation-guided mapping, the ASP-based aspects can be separated from the Java coding: the programmer does not even necessarily need to be aware of ASP.

Let us think of a user that designed (or has been given) a proper logic program $P$ to solve a sudoku puzzle and has also an initial schema. We assume that the initial schema is well-formed i.e. the complete schema solution exists and is unique. For instance, $P$ can correspond to the logic program presented in Section~\ref{subsec:KRR}, so that, coupled with a set of facts $F$ representing the given initial schema, allows to obtain the only admissible solution (i.e., a single answer set). It is worth remembering that, in case of less usual sudoku schemata featuring multiple solutions, the ASP program features multiple answer sets, one-to-one corresponding to such solutions.

By means of the annotation-guided mapping, the initial schema can be expressed in forms of Java objects. To this extent, we define the class \texttt{Cell}, aimed at representing a single cell of the sudoku schema, as follows:

\begin{lstlisting}[language=Java,
					basicstyle=\scriptsize\ttfamily,
					breaklines=true,
					tabsize=2,
					frame=TRBL,
					numbers=left,
					numbersep=1pt,
					numberblanklines=true,
					xleftmargin=2em,
					framexleftmargin=1.5em,
					keywordstyle=\bf]
@Predicate("cell")
public class Cell {

	@Term(1)
	private int row;
	
	@Term(2)
	private int column;
	
	@Term(3)
	private int value;
	
	[...]
	
}
\end{lstlisting}

It is worth noticing how the class has been annotated by two custom annotations, as introduced above. Thanks to these annotations the \texttt{ASPMapper} will be able to map \texttt{Cell} objects into strings properly recognizable from the ASP solver as logic facts of the form $cell(Row,Column,Value)$.

At this point, we can create an \texttt{Android Activity Component}~\footnote{\url{https://developer.android.com/reference/android/app/Activity.html}}, and start deploying our sudoku application:

\begin{lstlisting}[language=Java,
					basicstyle=\scriptsize\ttfamily,
					breaklines=true,
					tabsize=2,
					frame=TRBL,
					numbers=left,
					numbersep=1pt,
					numberblanklines=true,
					xleftmargin=2em,
					framexleftmargin=1.5em,
					keywordstyle=\bf]
public class MainActivity extends AppCompatActivity {
	
	[...]
	private Handler handler;

	@Override
	protected void onCreate(Bundle bundle) {
		handler = new AndroidHandler(getApplicationContext(),
	    	DLVAndroidService.class);
		[...]
	}
	
	public void onClick(final View view){
    [...]
		startReasoning();
	}

	public void startReasoning() {
		InputProgram inputProgram =
			new ASPInputProgram();
		for ( int i = 0; i < 9; i++){
			for ( int j = 0; j < 9; j++)
				try {
					if(sudokuMatrix[ i ] [ j ]!=0) {
						inputProgram.addObjectInput(
							new Cell(i, j, sudokuMatrix[i][j]));
					}
				} catch (Exception e) {	
					// Handle Exception
				}
		}
		handler.addProgram(inputProgram);

		String sudokuEncoding =
			getEncodingFromResources();			
		handler.addProgram(new
			ASPInputProgram(sudokuEncoding));

		Callback callback = new MyCallback();
		handler.startAsync(callback);
	}
}
\end{lstlisting}

The class contains a \texttt{Handler} instance as field, that is initialized when the \texttt{Activity} is created as an \texttt{AndroidHandler}.
Required parameters include the \texttt{Android Context} (an Android utility, needed to start an Android Service Component) and the type of \texttt{AndroidService} to use -- in our case, a \texttt{DLVAndroidService}.
In addition, in order to represent an initial sudoku schema, the class features a matrix of integers as another field where position $ (i,j) $ contains the value of cell $ (i,j) $ in the initial schema; cells initially empty are represented by positions containing zero.

The method \texttt{startReasoning} is in charge of actually managing the reasoning: in our case, it is invoked in response to a ``click'' event that is generated when the user asks for the solution.
Lines $19$--$32$ create an \texttt{InputProgram} object that is filled with \texttt{Cell} objects representing the initial schema, which is then served to the handler; lines $34$--$37$ provide it with the sudoku encoding.
It could be loaded, for instance, by means of a utility function that retrieves it from the \textit{Android Resources folder}, which, within Android applications, is typically meant for containing images, sounds, files and resources in general\footnote{\url{http://developer.android.com/guide/topics/resources}}.

At this point, the reasoning process can start; since for Android we provide only the asynchronous execution mode, a callback object is in charge of fetching the output when the ASP system has done (Lines $39$--$40$).

Eventually, once the computation is over, from within the callback function the output can be retrieved directly in form of Java objects. For instance, in our case an inner class \texttt{MyCallback} implements the interface \texttt{Callback}:

\begin{lstlisting}[language=Java,
					basicstyle=\scriptsize\ttfamily,
					breaklines=true,
					tabsize=2,
					frame=TRBL,
					numbers=left,
					numbersep=1pt,
					numberblanklines=true,
					xleftmargin=2em,
					framexleftmargin=1.5em,
					keywordstyle=\bf]
private class MyCallback implements Callback {

	@Override
	public void callback(Output o) {
		if(!(o instanceof AnswerSets))
			return;
		AnswerSets answerSets=(AnswerSets)o;
        if(answerSets.getAnswersets().isEmpty())
        	return;
        AnswerSet as = answerSets.getAnswersets().get(0);
		try {
			for(Object obj:as.getAtoms()) {
				Cell cell = (Cell) obj;
				sudokuMatrix[cell.getRow()]
					[cell.getColumn()] = cell.getValue();
			}
		} catch (Exception e) {
			// Handle Exception
		}
		displaySolution();
	}
}
 \end{lstlisting}

\subsection{Other Language Implementations of \embasp}
The implementation illustrated above relies on Java. Besides the fact that it represents a very popular, solid and reliable programming language, the choice was also motivated by the intention to foster the use of ASP in new scenarios, and in particular in the mobile one; Android is by far the most widespread mobile platform, and its development and deployment models heavily rely on Java.
However, as already stated, the abstract architecture of \embasp can be made concrete by means of other object-oriented programming languages. A thorough discussion about different language implementations is out of the scope of this work; however, we briefly discuss in the following some interesting possible approaches.

Most of components in the herein presented Java implementation have been accomplished thanks to features that are typical of any object-oriented language, such as \emph{inheritance} and \emph{polymorphism}.
The unique exception is represented by the \textit{ASPMapper} component, implemented by means of Java peculiar features, such as \emph{annotations} and \emph{reflection}.
In case of other languages that feature similar constructs, such as \texttt{C\#}~\footnote{Microsoft Developer Network, MSDN: C\# Attributes (\url{https://msdn.microsoft.com/en-us/library/mt653979}), C\# Reflection (\url{https://msdn.microsoft.com/en-us/library/mt656691})}, the approach can resemble the herein presented Java implementation.

With different languages that lack such features, the mapping mechanism can still be implemented with a simulation via inheritance and polymorphism and applying typical Software Engineering patterns~\cite{gamma1994design}.
As a matter of example, one possible implementation can be accomplished using the \textit{Prototype design pattern}, that results well-suited to our purposes, as it allows to ``specify the kinds of objects to create using a prototypical instance, and create new objects by copying this prototype''~\cite{gamma1994design}. Such pattern can be the key to simulate the dynamical loading of classes in languages that do not support it natively, as it happens with \texttt{C++}.
Indeed, the run-time environment can make use of it in order to automatically create an instance of each class when it's loaded, and then register the instance with a prototype manager -- in our case, represented by the \textit{ASPMapper} component.
All classes that in Java (or similar languages) would make use of reflection and annotations, can be defined by extending a properly defined \texttt{Prototype} class and then specify how to map predicates and terms. Moreover, a class \texttt{ASPMapper} would still be needed, with a behaviour quite similar to the Java case.

\section{ASP-based Applications: some Examples in the Educational Setting}
In this section we describe some ASP-based applications developed by means of \embasp for educational purposes, and, in particular, in the context of a university course that covers ASP topics; it is worth noting that such applications have been developed by some of the course attendants, i.e., undergraduate students. The educational aspect here is two-folded. The most relevant is the engagement of university (under)graduate students in ASP capabilities, in order to make them able to take advantage from it when solving problem and designing solutions, in the broadest sense. Furthermore, ASP looks well-fitted for the use in the development of educational/training software, as, for instance, the \emph{DLVEdu} app introduced below; a deeper study of such aspects, however, is out of the scope of the present work.

In the following, we first briefly introduce three applications; then, in order to further clarify the \embasp use, especially in the mobile setting, we describe the \dlvfit Android App more in detail.

\subsubsection*{GuessAndCheckers}
\emph{GuessAndCheckers} is a native mobile application that works as an helper for users that play ``live'' games of the (Italian) checkers (i.e., by means of physical board and pieces). The app, that runs on Android, can help a player at any time: by means of the device camera a picture of the board is taken, and the information about the current status of the game is properly inferred thanks to OpenCV\footnote{\url{http://opencv.org}}, an open source computer vision and machine learning software; an ASP-based artificial intelligence module then suggests the move.

Thanks to \embasp and the use of ASP, \emph{GuessAndCheckers} features a fully-declarative approach that made easy to develop and improve several different strategies, also experimenting with many combinations thereof.

The source code of this application along with the Android Application Package (APK) are available online; more details can be found at~\cite{embasp-web}.

\subsubsection*{DLVEdu}
\emph{DLVEdu} is an educational Android App for children, that integrates well-established mobile technologies, such as voice or drawn text recognition, with the modeling capabilities of ASP. In particular, it is able to guide the child throughout the learning tasks, by proposing a series of educational games, and developing a personalized educational path. The games are divided into four macro-areas: Logic, Numeric-Mathematical, Memory, and Verbal Language. The usage of ASP allows the application to adapt to the game experiences fulfilled by the user, her formative gap, and the obtained improvements.

The application continuously profiles the user by recording mistakes and successes, and dynamically builds and updates a customized educational path along the different games.

The application features a ``Parent Area'', that allows parents to monitor child's achievements and to express some preferences, such as desired express directions in order to grant/forbid access to some games or educational areas.

\subsubsection*{Connect4}
The popular turn-based \textit{Connect Four} game is played on a vertical 7*6 rectangular board, where two opponents drop their disks with the aim of creating a line of four, either horizontally, vertically, or diagonally.

The {\em Connect4} application allows a user to play the game (also known as \textit{Four-in-a-Row}) against an ASP-based artificial player. Notably, the declarative nature of ASP, its expressive power, and the possibility to compose programs by selecting proper rules, allowed to design and implement different AIs, ranging from the most powerful one, that implements advanced techniques for the perfect play, to the simplest one, that relies on some classical heuristic strategies. Furthermore, by using \embasp, two different versions of the same app have been built: one for Android, making use of DLV, and one for java-enabled desktop platforms, making use of clingo.

\subsubsection*{DLVfit}
The \dlvfit Android App was the first application making use of the framework; it was conceived as a proof of concept, in order to show the framework features and capabilities. To our knowledge, it is also the first mobile app natively running an ASP solver.

\dlvfit is a health app that aims at suggesting the owner of a mobile device the ``best'' way to achieve some fitness goals.
The app lets the user express her own goals and preferences in a very customizable ways along many combinable dimensions: calories to burn, time to spend, differentiation over several physical activities, time constraints, etc. Then, it monitors her actual activity throughout the day and, upon request, it computes one or more plans meant, if accomplished, to make her meet the aforementioned goals the way she would have preferred.

More in detail, the app constantly detects the current user activity (running, walking, cycling, etc.) and (at a customizable frequency) stores some information (activity type, timestamps, calories burned up to the present time, etc.). Activity detection is performed by means of the Google Activity Recognition APIs~\cite{google-activity-rec-web}, a de-facto standard on Android, thus relying on these for the accuracy of detection. As already mentioned, the user might ask, at any time, for a suggestion about a plan for the rest of the day; the reasoning module hence prepares a (set of) proper workout plans complying with the very personal goals and preferences previously expressed.

The user interacts with the app via a standard graphical interface; the reasoning module is actually in charge of building a proper ASP program, which is in turn fed to \dlv via \embasp. Such program matches the classical ``Guess/Check/Optimize'' paradigm introduced in Section~\ref{sec:ASP}, thus resulting easy to understand, enrich and customize:
\begin{itemize}
	\item
	the ``guess'' part chooses how much time to spend on each exercise;
	\item
	the ``check'' part forces the resulting plan to be admissible: burning the remaining amount of desired calories, do not exceed the time constraints, etc.;
	\item
	the ``optimize'' part, eventually, expresses preferences: minimize total time spent exercising, number of activities to perform, maximize the number of different activity types, avoid activities around a given time of the day, etc.
\end{itemize}

The logic program used takes as ``input'' (i.e., a set of facts as instances of proper predicates):
\begin{description}
	\item[calories\_burnt\_per\_activity(A, C)] \hfill \\
		the calories burnt (C), in each unit of time, per each Activity ($A$);
	\item[remaining\_calories\_to\_burn(R)] \hfill \\
		the remaining calories to burn in the rest of the current day;
	\item[how\_long(A, D)] \hfill \\
		the amount of time that can be spent for each activity $A$ (in order to reach the goal of burn all the remaining calories);
	\item[max\_time(T)] \hfill \\
		the duration of the workout (max: the remaining time to the end of day);
	\item[surplus(C)] \hfill \\
		the maximum surplus of calories to burn with the suggested workouts;
	\item[optimize(O, W, P)] \hfill \\
		the specific optimization operation(s) that the user wants to perform; each direction is assigned a weight ($W$) and a preference order ($P$).
\end{description}

An example of the basic input concepts described above is the following:

\medskip

\begin{lstlisting}[language=Prolog,frame=lines,label=code:input_example_1, keywordstyle=\linespread{1.1}\scriptsize\ttfamily, basicstyle=\linespread{1.1}\scriptsize\ttfamily]
calories_burnt_per_activity("ON_BICYCLE", 5).
calories_burnt_per_activity("WALKING", 2).
calories_burnt_per_activity("RUNNING", 11).

remaining_calories_to_burn(200).

how_long("ON_BICYCLE", 10).
how_long("ON_BICYCLE", 20).
how_long("WALKING", 10).
how_long("WALKING", 20).
how_long("RUNNING", 10).
how_long("RUNNING", 20).

max_time(20).

surplus(100).
\end{lstlisting}

\medskip

In this example the activities that can be performed (\texttt{"ON\_BICYCLE"}, \texttt{"WALKING"} and \texttt{"RUNNING"}) are specified along with the calories they allow to burn per unit of time; then, the amount of time spent for each activity is reported. Moreover, there are pieces of information about the calories that remain to burn in the current day (at least $200$, and up to $300$ due to the \texttt{surplus}) and the maximum time that the user wants to spend on the workouts ($20$).

Custom optimization preferences are typically represented as follows:

\medskip

\begin{lstlisting}[language=Prolog,frame=lines,label=code:input_example_2, keywordstyle=\linespread{1.1}\scriptsize\ttfamily, basicstyle=\linespread{1.1}\scriptsize\ttfamily]
optimize("RUNNING", 1, 3).
optimize("ON_BICYCLE", 3, 3).
optimize("WALKING", 2, 3).

optimize(time,0,2).

optimize(activities, 0, 1).
\end{lstlisting}

\medskip

Solutions, in this context, are actually workouts suggestions to the user. The \texttt{optimize} predicate is of arity $3$, and third argument is supposed to express the ``importance'' of the statement (the higher the number, the more the importance). In this example, the ASP code models that: $(i)$ the user wants  (preference level: $3$) to maximize the number of favourite activities to perform, and provides an order ( \texttt{"RUNNING"} first, then \texttt{"WALKING"} and finally \texttt{"ON\_BICYCLE"}); $(ii)$ if more than one admissible workout is found featuring the same favourite activities, she wants to minimize the total time spent exercising (preference level: $2$); also, $(iii)$ if there are workouts that have the same favourite activities and the same time, she wants to minimize the total number of activities (preference level: $1$).

The logic program is able to find the combinations of activities that should be performed in order to burn the remaining calories.
Obviously, this goal can be achieved, in general, in many different ways, each of them modelled by a different answer set.
Part of the rules of the program that we used are reported hereafter; full program is available online.

\medskip

\begin{lstlisting}[language=Prolog,frame=lines,label=code:logic_program, keywordstyle=\linespread{1.1}\scriptsize\ttfamily, basicstyle=\linespread{1.1}\scriptsize\ttfamily]
%%%%%% Guess Part %%%%%%
activity_to_do(A, HL) | not_activity_to_do(A, HL) :-
   how_long(A, HL).

%%%%%% Check Part %%%%%%
:- activity_to_do(A, HL1), activity_to_do(A, HL2),
   HL1 != HL2.

:- remaining_calories_to_burn(RC),
   total_calories_activity_to_do(CB), RC > CB.

:- remaining_calories_to_burn(RC),
   total_calories_activity_to_do(CB),
   CB > RCsurplus, RCsurplus = RC + surplus.

:- max_time(MTS), MTS < TS,
   total_time_activity_to_do(TS).

%%%%%% Optimize Part %%%%%%
:~ optimize(A, W, P), activity_to_do(A, _). [W:P]

:~ optimize(time, _, P), activity_to_do(_, HL). [HL:P]

:~ optimize(activities, _, P), #int(HM),
   #count{A, HL : activity_to_do(A, HL)} = HM. [HM:P]

\end{lstlisting}

\medskip

The \texttt{Guess Part} chooses how much time to spend on each exercise.
The \texttt{Check Part} checks that each activity selected has one specific amount of time, it ensures that all the remaining calories are burnt and that not more calories than the remaining (with the surplus) are burnt and it ensures to not exceed the maximum time that the user wants to spend on the workouts.
The \texttt{Optimize Part} makes use of \emph{weak constraints}\cite{bucc-etal-97a,calimeri2012asp}: in case the user specified preferences about activities, tries to select the favourite ones; in case she specified preferences about the time spent exercising, tries to minimize it; if she specified preferences about the number of different activities, tries to minimize it.

There is a wide range of customization possibilities in this setting: thanks to the modeling capabilities and the declarative nature of ASP, adding new features to \dlvfit, such as new exercises or new kind of preferences, is straightforward, and sums up to adding a few lines to the logic program.
It is also worth noting that the ASP program is dynamically built, thus providing the developer (and, in turn, the final user) with great customization and flexibility capabilities. Indeed, we plan to actually take advantage from this in the future versions of the prototype, contemplating a higher number of rules and sub-programs to be dynamically fed to \dlv.

\section{Related Work}
The problem of embedding ASP reasoning modules into external systems and/or externally controlling an ASP system has been already investigated in the literature; to our knowledge, the more widespread solutions are the \dlv Java Wrapper~\cite{ricc-2003}, JDLV~\cite{febb-etal-2012-kr}, Tweety~\cite{thimm2014tweety}, and the scripting facilities featured by clingo4~\cite{gekakasc14b}, which allow, to different extents, the interaction and the control of ASP solvers from external applications.

In clingo4, the scripting languages {\em lua} and {\em python} enable a form of control over the computational tasks of the embedded solver clingo, with the main purpose of supporting also dynamic and incremental reasoning; on the other hand, \embasp, similarly to the {\em Java Wrapper} and {\em JDLV}, acts like a versatile ``wrapper'' wherewith the developers can interact with the solver. However, differently from the Java Wrapper, \embasp features a \textit{Mapper} that, in the Java implementation, makes use of annotations, a form of metadata that can be examined at runtime, thus allowing an easy mapping of input/output to Java Objects; and differently from JDLV, that uses JPA annotations for defining how Java classes map to relations similarly to ORM frameworks, \embasp straightforwardly uses custom annotations, almost effortless to define, to deal with the mapping.

Moreover, our framework is not specifically bound to a single or specific solver; rather, it can be easily extended for dealing with different solvers; in addition, it allows to build applications that can run different solvers, and different instances, at the same time; none of the mentioned systems exposes this feature. Finally, to our knowledge, the specialization of \embasp for \dlv on Android has been the first actual attempt to port ASP solvers to mobile systems reported in literature; indeed, the preliminary version of \embasp was originally explicitly tailored to the mobile scenario~\cite{cali-etal-2015-aspocp,embasp-web}.

Tweety is an open source framework for experimenting on logical aspects of artificial intelligence; it consists of a set of Java libraries that allow to make use of several knowledge representation systems supporting different logic  formalisms, ranging from  classical  logics,  over  logic  programming  and  computational models for argumentation, to probabilistic modelling approaches, including ASP.
Tweety and \embasp cover a wide range of applications, and the use is very similar: at the bottom line, both provide libraries to incorporate proper calls to external declarative systems from within ``traditional'' applications. Currently, Tweety implementation is already very rich, covering a wide range of KR formalisms, yet looking less general, as the more abstract level is conceived as a coherent structure of Java libraries; also, it currently misses the mobile focus. \embasp is mainly focused on fostering the use of ASP in the widest range of contexts, as evidenced by the specialization for the mobile setting; nevertheless, the framework core is very abstract, and has been conceived in order to create libraries for different programming languages, platforms and formalisms.

\section{Conclusions}
In this paper we presented a framework for embedding ASP reasoning and modeling capabilities into external systems. The fully abstract architecture makes the framework general enough to be adapted to a wide range of scenarios; indeed, it can be implemented in any programming language, grounded to different platforms, and can make use of different ASP solvers. We herein presented an actual Java implementation and two specialized libraries for embedding DLV on Android applications and clingo on any Java-based desktop application.
The framework has been tested within some university courses featuring ASP topics, for implementing a set of applications, ranging from AI-based games to educative apps; it proved to be an effective set of tools and interoperability mechanisms able to ease the development of ASP-based applications, in both educational and real-world contexts.

It is worth noting that, although the framework has been mainly conceived for fostering the usage of ASP, its abstract core makes it also adaptable to other declarative knowledge representation formalisms.

\medskip

The framework, documentation, an application showcase and further details are freely available online~\cite{embasp-web}.

\section{Acknowledgements}
Francesco Calimeri has received funding from the European Union's Horizon 2020 research and innovation programme under the Marie Skodowska-Curie grant agreement No. 690974 for the project ``MIREL: MIning and REasoning with Legal texts''.

\bibliographystyle{abbrv}
\bibliography{ppdp-2016-help}

\end{document}